\documentclass[sigconf]{acmart}
\renewcommand\footnotetextcopyrightpermission[1]{} 
\usepackage{booktabs} 
\usepackage{soul}
\usepackage[ruled]{algorithm2e} 
\usepackage{tikz} 

\SetAlFnt{\small}
\SetAlCapFnt{\small}
\SetAlCapNameFnt{\small}
\SetAlCapHSkip{0pt}
\IncMargin{-\parindent}

\acmConference[VAMS'17]{The ACM RecSys Workshop on Value-Aware Multi-Stakeholder Recommendation}{August 27}{Como, Italy}

\setcopyright{acmcopyright}

\acmDOI{0000001.0000001_2}

\begin{document}
\title{Evaluating Music Recommender Systems for Groups} 


\author{Zsolt Mezei}
\affiliation{%
  \institution{Dept.\ of Computer Science}
  \streetaddress{ETH Z\"urich}
  \city{Z\"urich} 
  \country{Switzerland} 
}
\email{mezei3zsolt@yahoo.com}

\author{Carsten Eickhoff}
\affiliation{%
  \institution{Dept.\ of Computer Science}
  \streetaddress{ETH Z\"urich}
  \city{Z\"urich} 
  \country{Switzerland} 
}
\email{ecarsten@inf.ethz.ch}

\renewcommand{\shortauthors}{Zs.\ Mezei and C.\ Eickhoff}

\begin{abstract}

Recommendation to groups of users is a challenging and currently only passingly studied task. Especially the evaluation aspect often appears \textit{ad-hoc} and instead of truly evaluating on groups of users, synthesises groups by merging individual preferences. 

In this paper, we present a user study, recording the individual and shared preferences of actual groups of participants, resulting in a robust, standardized evaluation benchmark. Using this benchmarking dataset, that we share with the research community, we compare the respective performance of a wide range of music group recommendation techniques proposed in the literature. 
\end{abstract}

%
%

\ccsdesc[300]{Information systems~Recommender systems}
\ccsdesc[100]{Information systems~Personalization}
\ccsdesc[100]{General and reference~Surveys and overviews}
\ccsdesc[100]{General and reference~Evaluation}

%
%


\keywords{User study, evaluation, ranking, music, groups, recommendations.}

\maketitle


\section{Introduction}
\label{introduction}

Recommender systems are widely used both in industrial and academic settings. E-commerce platforms and content providers use them to recommend new products in order to maintain frequent interaction with their service. Most of these systems focus on individual users as the target of their recommendations (\textit{e.g.}, \cite{yahoo, internetradio, netflix, streamrec}). {\let\thefootnote\relax\footnote{Presented at the 2017 Workshop on Value-Aware and Multistakeholder Recommendation.}}

In turn, group recommendation is a new task in which an item is recommended to a group of users who will consume it together. Such a feature is useful for streaming movie or music providers when recommending content to a family or a group of friends who want to watch a film together. The same holds for search queries, travel offers or any type of products to be bought or consumed together. The challenge in such a task is modeling and matching the taste of the entire group even though only individual preferences may be known.
There are various alternative proposals in the academic literature regarding how to combine the preferences of individuals in some way to generate the group taste \cite{book, polylens, travel, web, approaches}.


As a consequence of the transient nature of groups, virtually all state-of-the-art music recommendation systems for groups evaluate their performance in an \textit{ad-hoc} manner on the basis of synthesised groups sampled from individuals that never actually interacted or formed a group. While this approach is easily implemented, it also na\"{i}vely ignores true group dynamics and makes for a sub-optimal benchmarking setup.

In this paper, we take a different approach by collecting true group preference data through a user study described in Section~\ref{userstudy}. To the best of our knowledge, this dataset represents the first resource of its kind. Finally, on the basis of our collection, Section~\ref{systemcomparison} compares a wide range of existing music group recommendation systems as well as basic machine learning algorithms.
\section{Survey Protocol}
\label{userstudy}

To collect a dataset of true expressions of group preferences, we conduct a supervised user study in which we team up participants in groups of three people, and, for a duration of approximately 15 minutes have them listen to music while engaging in a social activity (playing the card game Uno\footnote{https://en.wikipedia.org/wiki/Uno\_(card\_game)}). While doing so, they rate the various tracks that are played to them as a group, rather than individually.

The study is facilitated via a smartphone-centered Web application. After logging in to the study and providing basic demographic information (age and gender) the participants answer ten personality test questions\footnote{https://www.41q.com/} with two possible answer choices each. Finally, they are asked to select at least 5 favorite tracks out of a larger song database.

Once this basic information is recorded, the participants begin playing their card game while a shuffled random selection of 10 of their pooled favorite song choices is being played. In order to account for track length variability, each song was faded out after the first 1.5 minutes. At each song change, the mobile phone survey application prompts participants to individually rate the current track on a 5-point Likert scale, following the 1 (lowest) -- 5 (highest) star rating scheme employed by many popular content providers.

At the end of the experiment, the participants are asked to rate all played songs once more, as a group, needing to form an agreement between their individual preferences. During the entirety of the study, an experimenter is present in the room in order to take note of social interaction not recorded by the survey application. All subjects were recruited at and around a major university campus and were compensated for their participation. 
\section{Dataset Statistics}
\label{statistics}

We conducted 26 experiments, each involving three participants, resulting in an overall pool of 78 unique subjects. In total, 1068 user and 356 group ratings for songs were recorded. The average user rating is 3.36 with  standard deviation 1.27, while the average among groups is 3.3 with a tighter standard deviation of 1.08. Groups can be noted to be more reserved and choose the far ends of the rating scale less frequently than individual users.

\subsection{Songs}

The song database consists of 100 tracks and was populated with recently popular\footnote{http://www.billboard.com/charts/hot-100/2016-04-16}, as well as all-time favorite tracks\footnote{http://www.telegraph.co.uk/culture/music/11621427/best-songs-of-all-time.html}. An additional number of songs were added manually in order to increase diversity in terms of music styles.

Each track is annotated in terms of metadata information including the title, artists, album and genre. The list contains tracks by 87 different artists, 82 albums and 74 distinct genres. In case of 11 tracks information regarding the album is missing, and 4 tracks do not have genre tags.
Wave-form acoustic similarities between pairs of songs are computed using the Marsyas\footnote{http://marsyas.info/} sound analysis library as proposed by Sprague \textit{et al.}~\cite{partyvote}.

There are 17 songs that were never chosen by any participant of the study. On average an individual elected a song 3.99 times with a standard deviation of 3.98. Multiple users in a group could choose the same songs, on average a song is present 3.56 times in a group's playlist with standard deviation of 3.30.
The minimum size of a group's global playlist before sampling was 11 and the maximum 17, with an average of 13.69 and a standard deviation of 1.59.

\subsection{Qualitative Findings}

When asked for open-ended feedback after study completion, participants stated that they enjoyed the combination of a social activity and listening to music in order to make the study immersive. This frequently resulted in participants wanting to play out the last game of cards when the survey duration ended. No participant reported difficulties operating the user interface or switching contexts from playing the game to rating songs when tracks changed.

Participants enjoyed the group setup, and shared with the others sentiments like \textit{"That's my song!"} or \textit{"Who chose that one? It's terrible."} A few subjects had strong feelings linked to some of the songs, and it was important to them to influence the final group vote for those, while they did not care about most other tracks' ratings. Participant personality played a significant role in the observed interaction. While some of them were shy and did not actively join the discussion or debate, others were consistently more active/dominant. In one case, a subject individually gave 5 stars for a song, but during the group voting phase agreed when the group assigned only 3, without mentioning that he/she actually liked the track. When two subjects shared the same taste and rated songs in a similar manner, we noted them to suppress the third member's opinion in the discussion. During a few sessions, participants considered their performance in the game when rating tracks, giving higher votes during winning, rather than losing rounds.

Some of the groups agreed to rate tracks by assigning the average among individual votes as the group rating. Despite such prior agreements, several of these groups ended up discussing individual tracks instead of averaging. There was also a single case of a group not agreeing on how to vote, but after a long discussion finally settling for the average.

\section{System Comparison}
\label{systemcomparison}

This section compares a wide range of methods that attempt to infer group preferences from individual votes. The previous discussion of qualitative observations of group voting behavior underlines the non-trivial nature of this task even for humans. While previous work has neglected collecting actual group preferences and instead employs a range of approximations, here we use the observed actual group vote as ground truth. We propose two evaluation metrics to assess recommendation quality. Firstly, the mean squared error (MSE) between true and inferred ratings is calculated. In an alternative evaluation approach, casting the problem as a ranking task, we order songs according to decreasing inferred group preference and compute ranking quality in terms of nDCG. The relevance of each track is equal to the observed true group preference, ranging from 5 (most relevant) to 1 (least relevant)
Evaluation is performed in leave-one-out fashion, training models on 25 groups at a time and evaluating on the remaining one.

\subsection{Methods from Literature}

In this section, we present a comparison of 8 different algorithms from published papers. They were selected to represent the full spectrum of currently known group recommendation techniques and were implemented in Scala.

Ali \textit{et al}.~\cite{approaches} introduce three recommendation methods. Besides the geometric average as well as the minimum across all individual ratings, the authors discuss a weighted averaging policy, where the weight is given by the overall number of votes collected from the user, normalised by the total number of ratings. In our scenario, all group members have exactly the same number of votes. In consequence, we calculate weights on the basis of user attributes. We define a set of attributes \textit{A} such that the mean classification error is minimized. The weight is represented by the number of times a user's attribute is in the set \textit{A}.

Amer-Yahia \textit{et al}.~\cite{amer2009group} discuss a consensus function that, for user $u \in U$ and song $s$ discounts the average rating by the disagreement variance across individual ratings $r_{s,u}$ as presented in Equation \ref{disagreement}.

\begin{equation} \label{disagreement}
		r_{s} = \lambda avg_s + (1 - \lambda) (1- \frac { \sum_{u \in U} ( r_{s, u} - avg_{s} )^2 } {|U|})
\end{equation}

\textit{Dias et al}.~\cite{clustering} propose a  collaborative filtering approach based on latent factor spaces to recommend songs. They cluster users and choose leaders (the closest users to cluster centroids). These leaders' preferences are used to add diversity and smooth the group recommendations. The number of leading singular values were learned during the training process. The final score for a song is the weighted average of the group members and each leader. The mixture weight is the normalised number of ratings given by the user.

Chao \textit{et al}.~\cite{adaptiveradio} rely on negative ``preferences'' to recommend music. Such negative votes are defined by a rating $r_{s,u} \leq 3$. The system randomly plays any tracks which did not receive a negative vote from any group member, and is not on the same album with any song with a negative vote.

Liu \textit{et al}.~\cite{socialplaylist} play songs randomly from the group members' playlists. We generate random numbers from a continuous uniform distribution on the interval $[1, 5]$.

Kukka \textit{et al}.~\cite{ubirock}, O'Hara \textit{et al}.~\cite{jukola} and S\o{}rensen \textit{et al}.~\cite{meet} introduce different democratic voting methods. In each round the users can vote, and the highest voted song is played next. We approximate this behavior by the unweighted average across votes.

Sprague \textit{et al}.~\cite{partyvote} use democratic voting combined with wave-form song similarity. The final ranking criterion is a weighted mixture of the rating mean and the distance-discounted ratings of other songs. In this manner, explicit votes (0 distance to the current song) bear the greatest weight, while information from other, similar tracks can be elicited in the recommendation. 

\begin{equation} \label{partyvote}
		O_{song_i} = \lambda \cdot avg_{song_i} + \frac { (1 - \lambda) } {\#songs} \sum_{\substack{song_j \\ j \not= i}} (avg_{song_j} \cdot dist_{song_i, song_j})
\end{equation}

Table~\ref{papers} shows the performance comparison between the presented systems on our dataset. \cite{adaptiveradio} achieve low accuracy, as it merely divides songs into two sets: playable and not playable. The strongest overall performance, dependent on the chosen metric is observed for \cite{approaches}'s  unweighted averaging (lowest MSE) and \cite{partyvote}'s song similarity modification to this scheme (highest nDCG). We employ a Wilcoxon signed-rank test at $\alpha < 0.05$-level and find these leading methods to be statistically indistinguishable. The remaining field of methods shows significantly lower recommendation performance.

\begin{table}
\caption{Results of methods from academic literature}
	\centering
	\scalebox{0.83} {
		\begin{tabular}{l | c | c | c | c}
			\textbf{Method} & \textbf{MSE mean} & \textbf{MSE sd} & \textbf{nDCG mean} & \textbf{nDCG sd} \\ \hline
			\cite{approaches} avg & 0.593 & 0.299 & 0.942 & 0.048 \\ \hline
			\cite{approaches} min & 1.642 & 0.932 & 0.922 & 0.053 \\ \hline
			\cite{approaches} weighted & 0.637 & 0.344 & 0.937 & 0.059 \\	\hline
			\cite{amer2009group} & 0.596 & 0.299 & 0.941 & 0.052 \\ \hline
			\cite{clustering} & 0.635 & 0.284 & 0.938 & 0.049 \\	\hline
			\cite{adaptiveradio} & 1.316 & 0.459 & 0.865 & 0.030 \\ \hline
			\cite{socialplaylist} & 2.578 & 0.387 & 0.777 & 0.038 \\	\hline
			\cite{partyvote} & 0.586 & 0.299 & 0.938 & 0.053 \\ \hline
		\end{tabular}
	}
	\label{papers}
\end{table}

\subsection{Machine Learning Algorithms}

In addition to the domain-specific group recommendation techniques presented earlier, this section introduces a range of traditional machine learning as well as deep neural network techniques. All algorithms are implemented in Scala using the Spark Machine Learning Library\footnote{http://spark.apache.org/docs/latest/ml-guide.html} or Python and Tensorflow\footnote{https://www.tensorflow.org/}, respectively.


The classifiers attempt to predict observed group votes on the basis of a range of features: user attributes (demographics and personality questions) and individual ratings, average of all group members' votes, disagreement variance calculated in Equation \ref{disagreement}, minimum and maximum of the members' ratings, global average of all users' votes across songs, music metadata (title, artist, album, genre information), song similarity presented in~\cite{partyvote}.

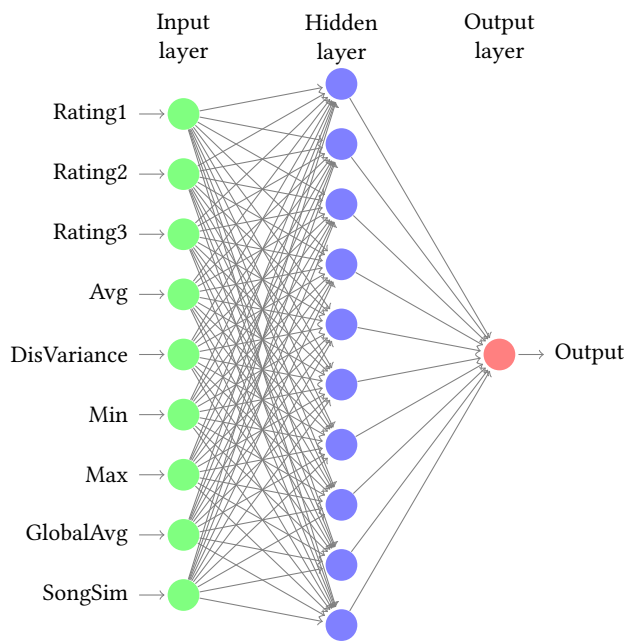
\begin{figure}
	\centering
	
    \def\layersep{2.1cm}

    \begin{tikzpicture}[shorten >=1pt,->,draw=black!50, node distance=\layersep]
        \tikzstyle{every pin edge}=[<-,shorten <=1pt]
        \tikzstyle{neuron}=[circle,fill=black!25,minimum size=12pt,inner sep=0pt]
        \tikzstyle{input neuron}=[neuron, fill=green!50];
        \tikzstyle{output neuron}=[neuron, fill=red!50];
        \tikzstyle{hidden neuron}=[neuron, fill=blue!50];
        \tikzstyle{annot} = [text width=4em, text centered]
    
        \node[input neuron, pin=left:Rating1] (I-1) at (0,-1) {};
        \node[input neuron, pin=left:Rating2] (I-2) at (0,-1.8) {};
        \node[input neuron, pin=left:Rating3] (I-3) at (0,-2.6) {};
        \node[input neuron, pin=left:Avg] (I-4) at (0,-3.4) {};
        \node[input neuron, pin=left:DisVariance] (I-5) at (0,-4.2) {};
        \node[input neuron, pin=left:Min] (I-6) at (0,-5.0) {};
        \node[input neuron, pin=left:Max] (I-7) at (0,-5.8) {};
        \node[input neuron, pin=left:GlobalAvg] (I-8) at (0,-6.6) {};
        \node[input neuron, pin=left:SongSim] (I-9) at (0,-7.4) {};
    
        \foreach \name / \y in {1,...,10}
            \node[hidden neuron] (H-\name) at (\layersep,0.2-0.8*\y) {};
    
        \node[output neuron,pin={[pin edge={->}]right:Output}] (O) at (2*\layersep,-4.2) {};
    
        \foreach \source in {1,...,9}
            \foreach \dest in {1,...,10}
                \path (I-\source) edge (H-\dest);
    
        \foreach \source in {1,...,10}
            \path (H-\source) edge (O);
    
        \node[annot,above of=H-1, node distance=0.6cm] (hl) {Hidden layer};
        \node[annot,left of=hl] {Input layer};
        \node[annot,right of=hl] {Output layer};
    \end{tikzpicture}

    \caption{Neural Network with 10 hidden units}
	\label{neuralnetwork}
\end{figure}

We investigated various feature subsets and combinations, finding linear regression to consistently show the lowest MSE, resulting in a performance comparable to that of the best previously described dedicated group recommendation schemes. Table~\ref{machine} gives a comprehensive overview of results. Using the same feature set, Figure~\ref{neuralnetwork} shows the architecture of the neural network with 10 hidden units, where the output is a rating from 1 to 5. All permutations of the group members' ratings (Rating1, Rating2, Rating3) are used during training.

\begin{figure}
	\centering
	\includegraphics[scale=0.35]{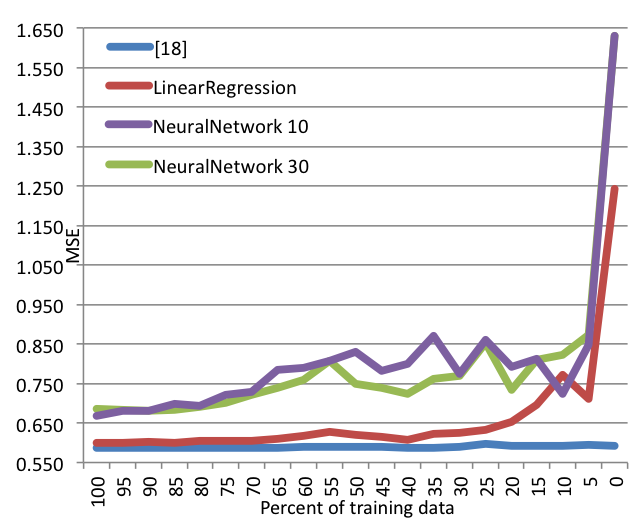}
	\caption{Recommendation performance of \cite{partyvote}, linear regression, neural network as a function of training set size.}
	\label{allMSE}
\end{figure}

Figure~\ref{allMSE} shows the results of \cite{partyvote}, a linear regression classifier with SGD Optimization, and a feed-forward neural network as we vary the amount of available training data.
The experiment suggests that further gains could be achieved if more training data had been available.
Linear regression and neural network algorithms show a more systematic benefit from additional data than did the group recommendation schemes analyzed before. Although some configurations of these supervised classifiers were able to match or mildly outperform the previously strongest methods, the limited dataset size did not allow us to confirm statistical significance of such differences. 

\begin{table}
\caption{Results of machine learning algorithms}
	\centering
	\scalebox{0.83} {
		\begin{tabular}{l | c | c | c | c}
			\textbf{Method} & \textbf{MSE mean} & \textbf{MSE sd} & \textbf{nDCG mean} & \textbf{nDCG sd} \\ \hline
			LinearRegression    & 0.599 & 0.306 & 0.942 & 0.056 \\ \hline
			Lasso               & 1.194 & 0.393 & 0.851 & 0.051 \\ \hline
			RidgeRegression     & 0.779 & 0.307 & 0.935 & 0.052 \\ \hline
			LogisticRegression  & 0.881 & 0.393 & 0.889 & 0.053 \\ \hline
			DecisionTreeClas    & 0.782 & 0.383 & 0.910 & 0.052 \\ \hline
			DecisionTreeReg     & 0.623 & 0.295 & 0.938 & 0.050 \\ \hline
			NaiveBayes          & 1.312 & 0.653 & 0.802 & 0.040 \\ \hline
			RandomForestClas    & 0.663 & 0.312 & 0.919 & 0.056 \\ \hline
			RandomForestReg     & 0.612 & 0.286 & 0.944 & 0.044 \\ \hline
			GradientTreeReg     & 0.626 & 0.297 & 0.937 & 0.050 \\ \hline
			NeuralNetwork 10    & 0.668 & 0.274 & 0.910 & 0.057 \\ \hline
			NeuralNetwork 30    & 0.686 & 0.285 & 0.903 & 0.063 \\ \hline
			NeuralNetwork 60,40 & 0.670 & 0.299 & 0.922 & 0.051 \\ \hline
		\end{tabular}
	}
	
	\label{machine}
\end{table}

\section{Conclusion}
\label{conclusion}

In this paper, we describe a user study and benchmarking dataset in which natural interactions of 78 participants with music recommendation systems for groups are elicited. The goal of this study is to provide a robust testbed for evaluation of group recommendation systems. Demonstrating the merit of this resource, we evaluate a wide range of state-of-the-art music recommendation systems for groups as well as general-purpose machine learning methods on the task of inferring group preferences from individual user votes.

Among the compared methods, despite the presence of a few local optima, unweighted averaging of individual votes has been confirmed as the most robust and generally applicable choice of recommendation scheme.


This study investigated fixed group sizes of three participants in order to limit inter-experiment variance. In the future, it would be interesting to study different group sizes as well as the behavior of organic (\textit{e.g.}, friends, families, etc.) versus randomly assigned groups. On the recommender systems side, it would be interesting to investigate playlist generation for groups (rather than recommending individual tracks as studied here). Besides meeting the group's preferences, such playlists need to satisfy constraints such as continuity, diversity or  similarity~\cite{playlistgeneration}.

%
%
%

\newpage
\bibliographystyle{ACM-Reference-Format}
\bibliography{ref.bib}

\end{document}